%% file: eccv2020submission.tex
\documentclass[runningheads]{llncs}
\usepackage{graphicx}
\usepackage{comment}
\usepackage{amsmath,amssymb} 
\usepackage{color}
\usepackage{epsfig}
\usepackage{svg}
\usepackage{amsmath,bm}
\usepackage{amssymb}
\usepackage{multirow}
\usepackage{siunitx}
\usepackage[width=122mm,left=12mm,paperwidth=146mm,height=193mm,top=12mm,paperheight=217mm]{geometry}

\usepackage[perpage]{manyfoot}
\DeclareNewFootnote{B}[fnsymbol]

\begin{document}
\pagestyle{headings}
\mainmatter
\def\ECCVSubNumber{5050}  

\title{Learning Inverse Rendering of Faces from Real-world Videos} 

\titlerunning{Inverse Face Rendering}
%
\author{
Yuda Qiu\inst{\footnotemarkB[4]1,2} \and
Zhangyang Xiong\inst{\footnotemarkB[4]1,3}\and
Kai Han\inst{4} \and
Zhongyuan Wang\inst{3} \and \\
Zixiang Xiong\inst{5} \and
Xiaoguang Han \inst{\footnotemarkB[1]1,2}
}
\authorrunning{Inverse Face Rendering}
%
\institute{
Shenzhen Research Inst. of Big Data \and
The Chinese University of Hong Kong, Shenzhen\\
\inst{3}Wuhan University \quad
\inst{4}University of Oxford \quad
\inst{5}Texas A\&M University 
}

\footnotetextB[4]{Equal contribution.}
\footnotetextB[1]{Corresponding author: \email{hanxiaoguang@cuhk.edu.cn}}


\maketitle
\begin{abstract}
In this paper we examine the problem of inverse rendering of real face images. Existing methods decompose a face image into three components (albedo, normal, and illumination) by supervised training on synthetic face data. However, due to the domain gap between real and synthetic face images, a model trained on synthetic data often does not generalize well to real data. Meanwhile, since no ground truth for any component is available for real images, it is not feasible to conduct supervised learning on real face images. To alleviate this problem, we propose a weakly supervised training approach to train our model on real face videos, based on the assumption of consistency of albedo and normal across different frames, thus bridging the gap between real and synthetic face images. In addition, we introduce a learning framework, called IlluRes-SfSNet, to further extract the residual map to capture the global illumination effects that give the fine details that are largely ignored in existing methods. Our network is trained on both real and synthetic data, benefiting from both. We comprehensively evaluate our methods on various benchmarks, obtaining better inverse rendering results than the state-of-the-art.

\end{abstract}
\begin{figure}[htb]
\fontsize{8}{10}\selectfont
  \centering
  \includegraphics[width=3.2in]{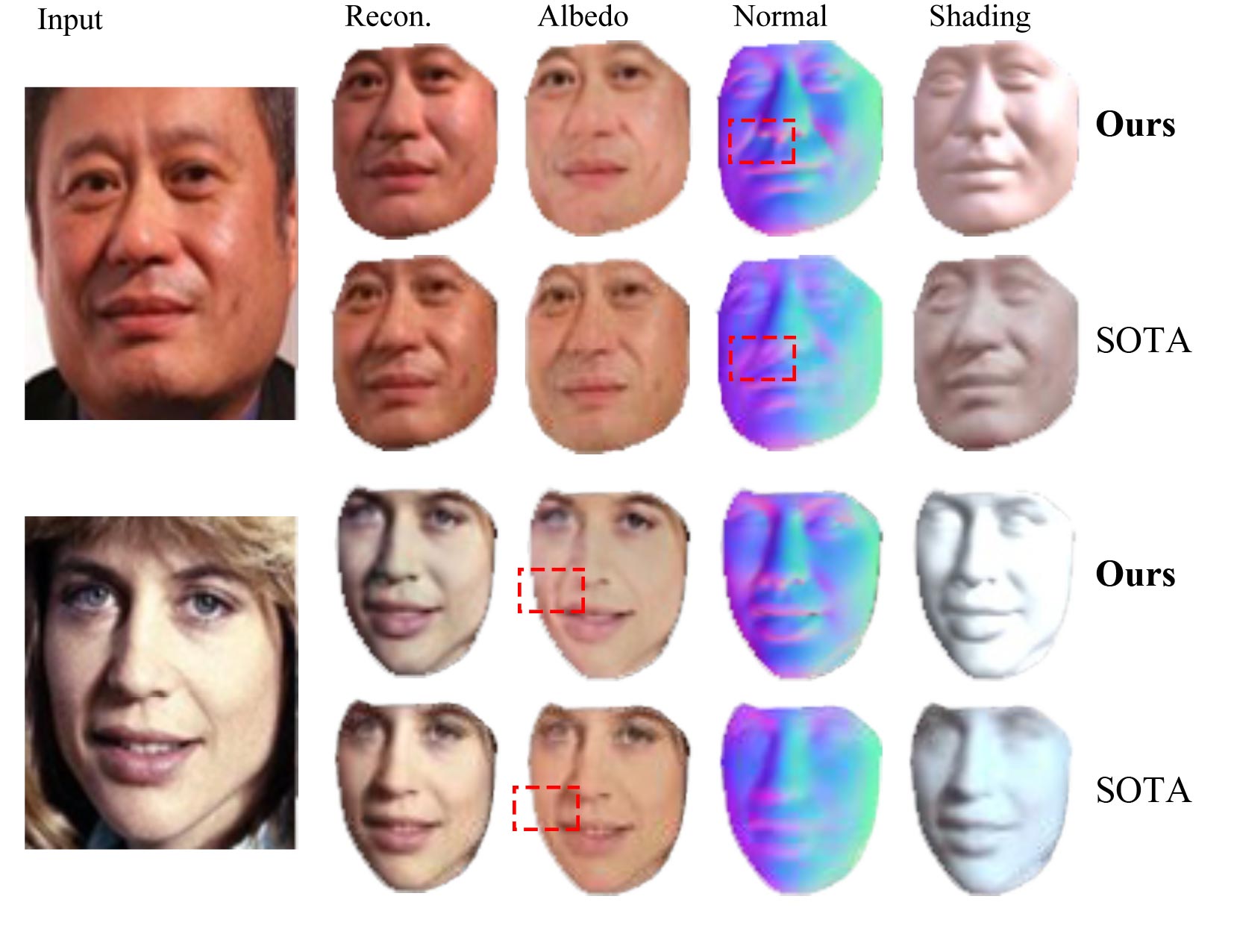}
  \caption{Decomposing real world faces into albedo, normal, and illumination. The first row of each sample shows our results, and the second one is SfSNet,\cite{sengupta2018sfsnet}, the state-of-the-art work on inverse face rendering. (Best viewed in PDF with zoom.)}
\end{figure}

\input{intro_ver2}
\input{related_work}
\input{method}

\input{experiment}

\input{application}

\input{conclusion}

\section*{Acknowledgment}
\noindent The work was supported in part by grants No. 2018YFB1800800, No. 2018B030\\338001, No. 2017ZT0 7X152, No. ZDSYS201707251409055 and in part by National Natural Science Foundation of China (Grant No.: 61902334 and 61629101).  
%
%
\newpage
\bibliographystyle{splncs04}
\bibliography{egbib}
\input{Appendix}

\end{document}

%% file: intro_ver2.tex
\section{Introduction}

Inverse rendering aims at estimating the components of an image in its formation process.
An image is often decomposed to three components, namely, albedo (reflectance properties), normal (shape attributes), and illumination \cite{aldrian2012inverse}, \cite{barron2014shape}, \cite{shu2017neural}, \cite{yu2019inverserendernet}.
Inverse rendering has important applications in image analysis (e.g., scene segmentation and material recognition) and editing (e.g., photo relighting).

In this paper, we consider inverse rendering of human face images because they belong to the most important class of images in vision and recognition tasks. Promising results have been achieved under constrained scenarios (e.g., on the well designed lighting stages\cite{debevec2000acquiring}\cite{bradley2010high}\cite{ghosh2011multiview}). However, inverse rendering of in-the-wild face images remains an open problem due to complex variations in human face appearances, illumination conditions, and shadows. Moreover,
the lack of ground-truth decomposition components makes this task highly ill-posed.
To tackle this problem,
some hand-crafted priors (e.g., \cite{kimmel2003variational}, \cite{barron2014shape}) have been introduced for each component to guide the decomposition. Unfortunately, these priors make strong assumptions on face attributes (e.g., strong shape and reflectance priors for Caucasians) that are tailored for specific tasks. Thus they do not generalize well to in-the-wild face images \cite{bonneel2017intrinsic}.

Owing to the success of deep learning in many tasks, deep convolutional neural networks (CNNs) have recently been employed to address the problem of inverse rendering. For example, \cite{shu2017neural,sengupta2018sfsnet} employed CNNs to deal with this problem by training on synthetic data.
After training on synthetic face image datasets, the network is applied on real face images to obtain different image modeling components.
However, training only on synthetic data does not generalize well to real data since real face images contain much richer facial variations that cannot be captured by synthetic data (e.g., faces wearing glasses and/or makeup, faces with beard, etc).
It is thus very desirable to make use of real data for training.
To this end, \cite{sengupta2018sfsnet} proposed to train on real face images with pseudo labels.
However, results reported in \cite{sengupta2018sfsnet} are far from satisfactory because the pseudo labels obtained from a network pre-trained on synthetic data cannot model complex variations in real face images.

Instead of using static independent real images, images from different viewpoints or video sequences of the same scene have been used to reduce the gap between real and synthetic data for intrinsic decomposition of indoor and/or outdoor scenes (e.g., \cite{Kim2016MultiviewIR}, \cite{yu2019inverserendernet}, \cite{Li2018LearningII}).

Inspired by the above approach, in this work we consider the use of images from real face videos\footnote{We do not employ multi-view images because they do not provide additional lighting constraints.} to bridge the domain gap because consistency constraints can be derived from different frames of the same person and they provide a stronger error measure in learning than the simple image reconstruction loss.
Our approach is motivated by the key observation that the albedo and normal maps of two face frames in a same-person video only differ in pose and expression and that consistency of these maps can be derived. Specifically, we transform the components between frames to algin albedo and normal: the transformation of albedo is the displacement of corresponding pixels, while for normal it consists of displacement and change of directions.
Note that such transformations cannot be achieved by traditional (e.g. optical flow) algorithm due to the texture-less nature of albedo and directional shifts in normal.
After aligning one frame with another, we leverage the consistency constraint between their albedo and normal maps to regularize the decomposition procedure, leading to weakly supervised learning on real face images.
This way, the domain gap between real and synthetic face images is drastically reduced. Our experiments corroborates that the alignment (though not perfect) obtained by \emph{AlignNet$_a$} and \emph{AlignNet$_n$} can effectively improve the inverse rendering performance on real data.

We also note that existing methods do not take into account high frequency details such as shadow and highlights, making the inverse rendering results less realistic.
Instead of considering the inverse rendering problem as decomposing an image into three components, we treat it as decomposing an image into four components (albedo, normal, illumination, and residual).
The advantage of the additional residual map component in our new approach is that it leads  to more realistic face images that maintain high frequency details induced by global illumination effects such as shadow and highlights.
We propose a learning framework called IlluRes-SfSNet to perform face image decomposition (into albedo, normal, illumination, and residual).
We train IlluRes-SfSNet on both real and synthetic data\footnote{We create two synthetic datasets (in addition to an existing one) to train our model.}, benefiting from both, and compute the albedo, normal, illumination components in a similar way to existing works (e.g., the SfSNet \cite{sengupta2018sfsnet}). The residual map is obtained by subtracting the image with global illumination and the one that only contains local illumination.
We run extensive experiments to evaluate IlluRes-SfSNet on various benchmarks, obtaining much better results than the state-of-the-art.




To summarize, our main contributions are threefold:
\begin{itemize}
   \item The conceptual significance of introducing a consistency assumption of albedo and normal of the same person in a video. This consistency assumption led to the idea of weakly supervised training of our neural network model, hence bridging the domain gap between synthetic and real images.

    \item Our proposed dual CNN models \emph{AlignNet$_a$} and \emph{AlignNet$_n$} as workhorses for learning the alignment between albedo and normal maps of different frames in a same-person face video. After pretraining on synthetic data, they are applied to real data to align the predicted albedo and normal maps, enabling weakly supervised learning on real data.
        
    \item An IlluRes-SfSnet learning framework that seamlessly integrates SfSnet
    \cite{sengupta2018sfsnet} and Illumination Residual Net to predict a residual map, in addition to albedo, normal, and illumination, for better inverse rendering results than the state-of-the-art.
    
\end{itemize}

Our current work opens doors to development of exciting applications such as relighting and albedo editing, for which we show some results in the end of the paper. 

%% file: related_work.tex
\section{Related work}

\noindent \textbf{Inverse rendering of images:} Decomposing an image into its intrinsic components is a long-standing and challenging task in computer vision.
The most popular forms are intrinsic images \cite{horn1974determining}\cite{barrow1978recovering}\cite{tappen2003recovering}\cite{bonneel2017intrinsic}, which define the decomposition layers as reflectance and shading (the function of shape and illumination).
Recently, SIRFS \cite{barron2014shape} showed that further recovering surface normal and lighting from shading can improve the performance of decomposition.
Since it is impossible to learn the decomposition without any constraints on the intrinsic components,
classical algorithms for inverse rendering usually rely on a sophisticated design of priors, such as sparsity of reflectance \cite{shen2011intrinsic}\cite{rother2011recovering}, user strokes \cite{bousseau2009user}\cite{shen2013intrinsic}, and RGB-D settings \cite{lee2012estimation}\cite{barron2013intrinsic}\cite{chen2013simple}.
These priors lead to promising decomposition in specific applications but do not generalize well to in-the-wild images.
Deep learning has also been applied to inverse rendering.
Given the lack of real ground truth data, some works perform supervised learning on synthetic datasets\cite{narihira2015direct}\cite{fan2018revisiting}\cite{lettry2018darn}, but they still suffer from poor performance on real data. Our paper focuses on inverse rendering of human face images, which commonly serve as major objects in real world photos.

\noindent \textbf{Inverse rendering of human face:} Since 3D morphable model (3DMM) of faces was proposed in 1999 \cite{blanz1999morphable}, it has served as a statistical shape prior for face inverse rendering \cite{lee2005estimation}\cite{wang2008face}\cite{kemelmacher20103d}\cite{li2014intrinsic}.
Tewari {\em et al.} \cite{tewari2017mofa} combined deep learning- and model-based capture in an end-to-end network to infer intrinsic components from a single input image, with a well designed differential parametric decoder;
they \cite{tewari2019fml} further developed an algorithm to learn the facial shape and reflectance variations from uncontrained images, without a pre-existing shape identity or albedo model.
However, these works are still based on parametric models on shape and reflectance, which resulted in loss of details.

Another branch of research focuses on model-free structures \cite{shu2017neural}\cite{sengupta2018sfsnet}.
Sengupta {\em et al.} \cite{sengupta2018sfsnet} designed a training paradigm called SfSNet to learn fine-scale separation of albedo and normal;
they first trained a simple network on synthetic data to obtain coarse estimations for real images and then trained another decomposition architecture with residual blocks, on both synthetic and real data;
they claim that SfSNet can outperform state-of-the-art algorithms for inverse rendering of faces, but SfSNet's performance is restricted by the coarsely estimated labels of real data.
In addition, SfSNet only models local illumination of images, ignoring the spatially-varying components. This leads to artifacts on albedo and normal map, when there are obvious cast shadows or mutual illumination.
We propose a weakly supervised learning on real face videos, avoiding the dependence on labels of real data. Moreover, we design an IlluRes-SfSNet framework to extract spatially-varying illumination information.  


\noindent \textbf{Inverse rendering of in-the-wild images:} Due to the lack of dataset, multi-image based approaches have been introduced, with consistency of scene variables in images being used to constrain the solution, especially the similarity in albedo of the same object.
Researchers in \cite{laffont2015intrinsic}\cite{ma2018single}\cite{Li2018LearningII} trained their network with a set of images of a scene under varying illuminations but the same viewpoint, to help disambiguate albedo and shading, and the author of \cite{yu2019inverserendernet} further removed the constraint on the viewpoint of inputs by warping albedo before measuring similarity.


%% file: method.tex
\section{Method}
In this section, we introduce our method for inverse rendering of in-the-wild face images. Similar to~\cite{sengupta2018sfsnet}, we also consider human faces as Lambertian surfaces. Given the albedo $\mathbf{A}_{p\times q \times 3}$, normal $\mathbf{N}_{p\times q \times 3}$, and lighting $\mathbf{L}_{9 \times 3}$\footnote{Following~\cite{sengupta2018sfsnet}, we also use the first-three order spherical harmonic coefficients to encode lighting, and we repeat three times for each color channel here.}
, the face image $\mathbf{I}_{p\times q \times 3}$ can be rendered by

\begin{equation}
\mathbf{I} = \mathbf{A} \circ f(\mathbf{N}, \mathbf{L}) ,
\label{sfsdefine}
\end{equation}
where $f(, )$ denotes the function that renders the shading image from normal and lighting, and $\circ$ denotes element-wise multiplication.

However, the above image formation model does not consider global illumination effects such as shadow, highlights, light interactions, etc\footnote{Here we follow the same definitions as in \cite{schneider2017efficient} for global illumination and local illumination.}. Thus, the face images rendered using the above model are often less realistic. In order to alleviate this problem, we propose to add a residual map to account for the global illumination effects. The enhanced image formation model can then be written as

\begin{equation}
\mathbf{I}_g = \mathbf{I}_l + \mathbf{R} ,
\label{ourdefine}
\end{equation}
where $\mathbf{I}_g$ denotes the image considering global illumination, $\mathbf{I}_l = \mathbf{A} \circ f(\mathbf{N}, \mathbf{L})$ stands for the image only considering local illumination, and $\mathbf{R}$ is the residual map that compensates global illumination effects.

Our objective is to decompose the face images into $\mathbf{A}$, $\mathbf{N}$, $\mathbf{L}$, and $\mathbf{R}$, which can be used to render realistic face images. To achieve this goal, we first make use of synthetic data. We create two synthetic datasets and expand one dataset from existing synthetic datasets. Each data sample contains $\mathbf{I}_g$, $\mathbf{I}_l$, $\mathbf{A}$, $\mathbf{N}$, and $\mathbf{L}$, allowing fully supervised learning. However, training on synthetic data alone is not enough to decompose real face images faithfully, since there is a huge domain gap between real and synthetic images. Therefore, it is essential to include real face data in our training. Unfortunately, there is no ground-truth $\mathbf{I}_l$, $\mathbf{A}$, $\mathbf{N}$, and $\mathbf{L}$ available for real face images. To get around the problem, we propose a weakly supervised learning method using real face videos.

In the sequel, we cover supervised learning in IlluRes-SfSNet in Section~\ref{sec:sup} first, we then describe weakly supervised learning by further introducing \emph{AlignNet$_a$} and \emph{AlignNet$_n$} in Section \ref{sec:weak_sup}.

\begin{figure}[htb]\centering
\fontsize{5}{10}\selectfont
\includegraphics[width=3.5in]{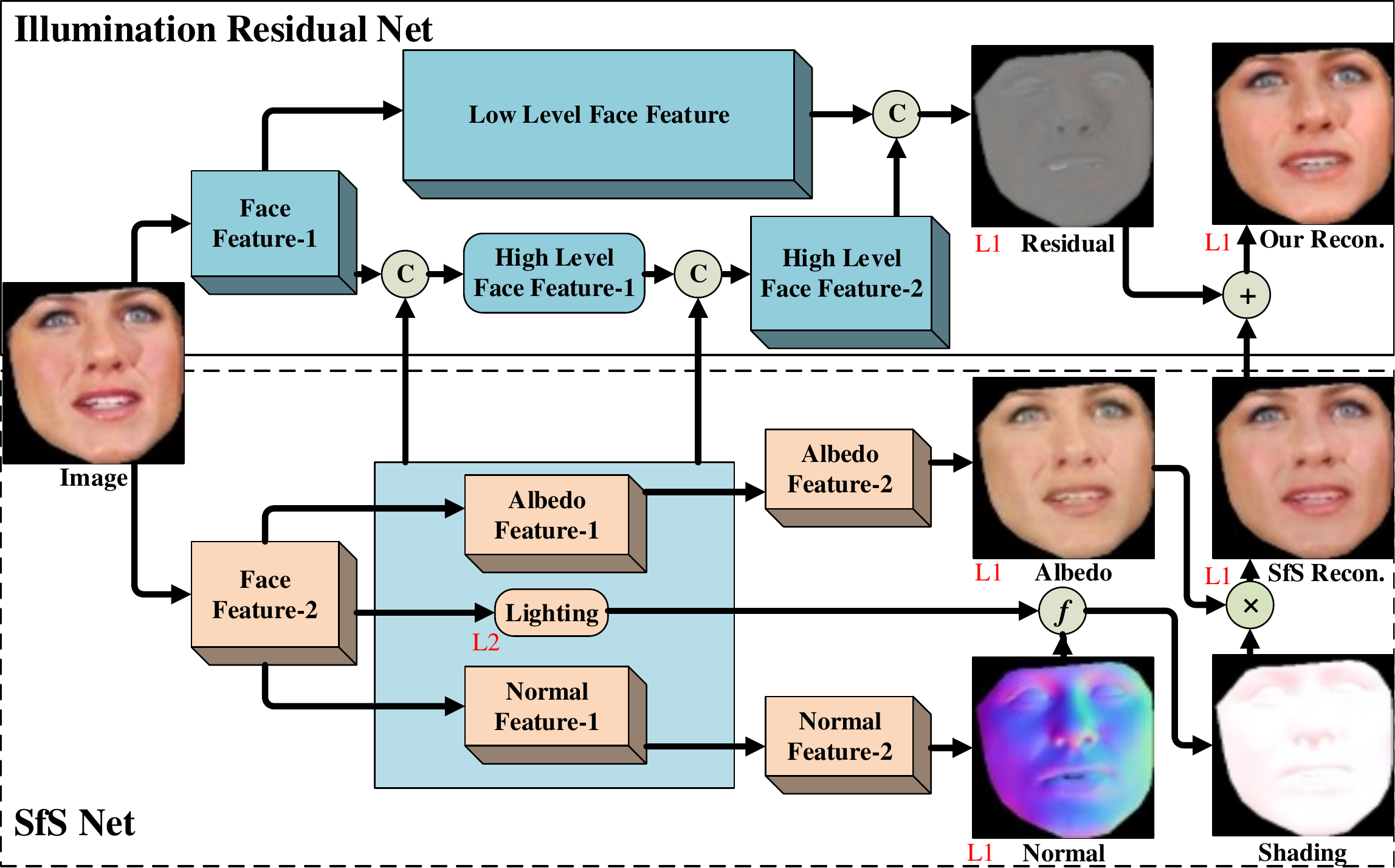}
\caption{The IlluRes-SfSNet architecture that decomposes an image into albedo, normal, lighting, and residual. It consists of two subnets: SfSNet and Illumination Residual Net, with SfSNet decomposing a face image into albedo, normal, and lighting, and Illumination Residual Net predicting the residual map that accounts for the global illumination effects such as shadow, highlights, etc. The two subnets are seamlessly combined with each other.}\label{Network Archtecture}
\end{figure}

\subsection{Supervised learning on synthetic data}
\label{sec:sup}

Our learning framework, called IlluRes-SfSNet, for supervised learning on synthetic data is shown in Fig.\ref{Network Archtecture}. In general, our model takes a single face image as input and decomposes it into its corresponding albedo, normal, lighting, and residual map.
IlluRes-SfSNet consists of two subnets: SfSNet for decomposing a face image into normal map $\mathbf{N}$, albedo $\mathbf{A}$, and lighting $\mathbf{L}$, and Illumination Residual Net for extracting the global illumination effects to compensate details that cannot be obtained by SfSNet.
The final face image $\mathbf{I}_g$ is rendered by adding the residual map $\mathbf{R}$ from Illumination Residual Net to the face image $\mathbf{I}_l$ with local illumination obtained from SfSNet.
Note that the two subnets are seamlessly combined with each other: features learned in SfSNet are combined with those learned in Illumination Residual Net for better residual map estimation, and the gradients for learning $\mathbf{I}_g$ are back-propagated through both SfSNet and Illumination Residual Net.
In this manner, information learned by both subnets mutually enhance each other.
IlluRes-SfSNet is trained in two stages on synthetic data. In the first stage, we train SfSNet by minimizing
\begin{equation}
L_s = \lambda_l\|\mathbf{I}_l - \bar{\mathbf{I}}_l \| +  \lambda_a\|\mathbf{A} - \bar{\mathbf{A}} \| + \lambda_n\|\mathbf{N} - \bar{\mathbf{N}} \| + \lambda_h\|\mathbf{L} - \bar{\mathbf{L}} \|^2,
\label{loss_synth_local}
\end{equation}
where $\mathbf{X}$ and $\bar{\mathbf{X}}$ represent the ground truth and the prediction, respectively, and $\lambda_{\{l, a, n, h\}}$ the weights of each loss term. In the second stage, we train the whole IlluRes-SfSNet with the loss function
\begin{equation}
L_g = \lambda_g\|\mathbf{I}_g - \bar{\mathbf{I}}_g \| +  \lambda_r\|\mathbf{R} - \bar{\mathbf{R}} \| + L_s,
\label{loss_synth_global}
\end{equation}
where $\lambda_g$ and $\lambda_r$ represent the weights for image reconstruction and residual regression losses.

\subsection{Weakly supervised learning on real data}
\label{sec:weak_sup}

As mentioned earlier, although our IlluRes-SfSNet can improve the details by considering global illumination effects, training on synthetic data alone is still not enough to bridge the gap between real and synthetic data. Due to the lack of ground-truth decomposition in real image datasets, it is not possible to directly train on real face images. Simply adopting reconstruction loss on real images does not help, since reconstruction loss alone does not provide any constraint on decomposition of real face images.
Instead, we propose to train on real face videos to mitigate this problem. Although the ground truth is still not available, we can subtly make use of the consistency among video frames of the same identity, thus allowing weakly supervised learning on real face data.

Specifically, we observe that two different frames in the face video of the same identity only differ in pose and expression, which indicates that albedo/normal maps between them are subject to a transformation. However, this is not the case for lighting, since the shadings between different frames may vary a lot.
The transformation between different albedo/normal maps can be easily learned using fully supervised method~\cite{zakharov2019few}. Since albedo and normal are independent from lighting and lighting in real data is much more complicated than synthetic data, we conjecture that the domain gap between albedo and normal of synthetic and real data is smaller than that between images of synthetic and real images.
Therefore, we propose to learn the transformation between albedo and normal of different frames from the same person using synthetic data; this way the trained model can be reliably transferred to real images.

Note that, similar to SfSNet \cite{sengupta2018sfsnet}, IlluRes-SfSNet is model free. However, the key distinction between them is that there is no need for coarsely estimated labels of real data in IlluRes-SfSNet. Instead, it exploits consistency of albedo and normal in synthetic and real data.



Instead of using a single CNN to align albedo and normal maps of different face images, we use two CNNs, namely, \emph{AlignNet$_a$} and \emph{AlignNet$_n$}, for albedo and normal, respectively, to account for context information.
In particular, the transformation between albedo of different images of the same person is mainly geometric transformation.
Despite of geometric transformation, normal directions of the same point will change with different poses and expressions.
This explains why we adopt two CNNs instead of one.

To enforce the network to learn the geometric transformations for albedo and normal, we propose to train \emph{AlignNet$_a$} and \emph{AlignNet$_n$} using face contours, which mainly contain the geometric transformation, together with the albedo and normal maps. Using albedo and normal maps alone may result in trivial solutions such as color transformation. The face contour $\mathbf{C}$ can be easily obtained by detecting facial landmarks on a face image and connecting the detected landmarks on each parts such as eyes, nose, and mouth \cite{Faceplusplus}.

\begin{figure} \centering
\includegraphics[width=3.5in]{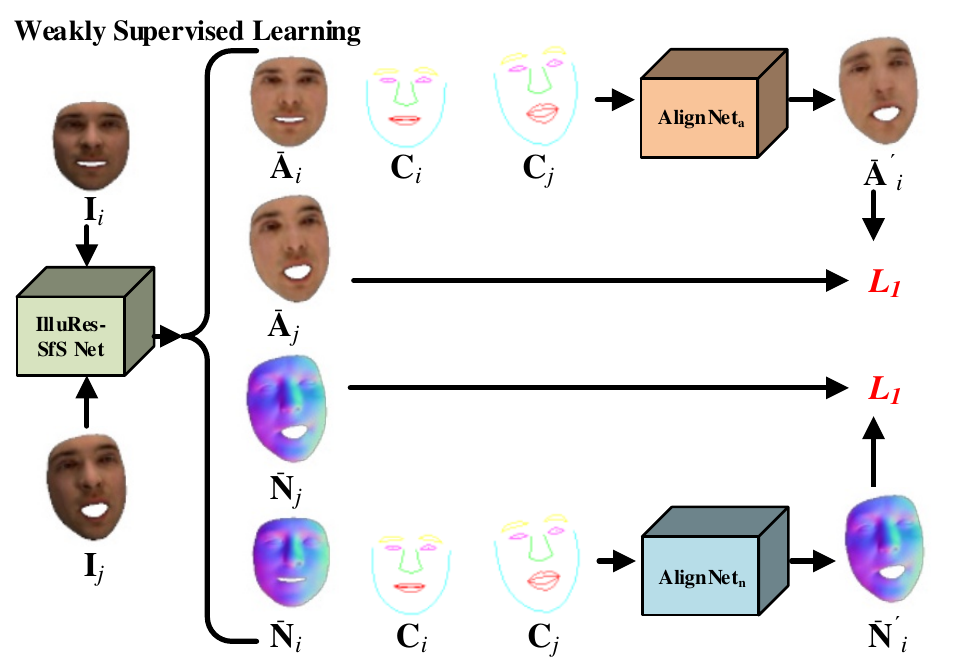}
\caption{Weakly supervised learning of $AlignNet_a$ and $AlignNet_n$ in IlluRes-SfSNet on real video data.}
\label{weakly_supervised_learning} \end{figure}

\begin{figure} \centering
\fontsize{8}{10}\selectfont
\includegraphics[width=4.8in]{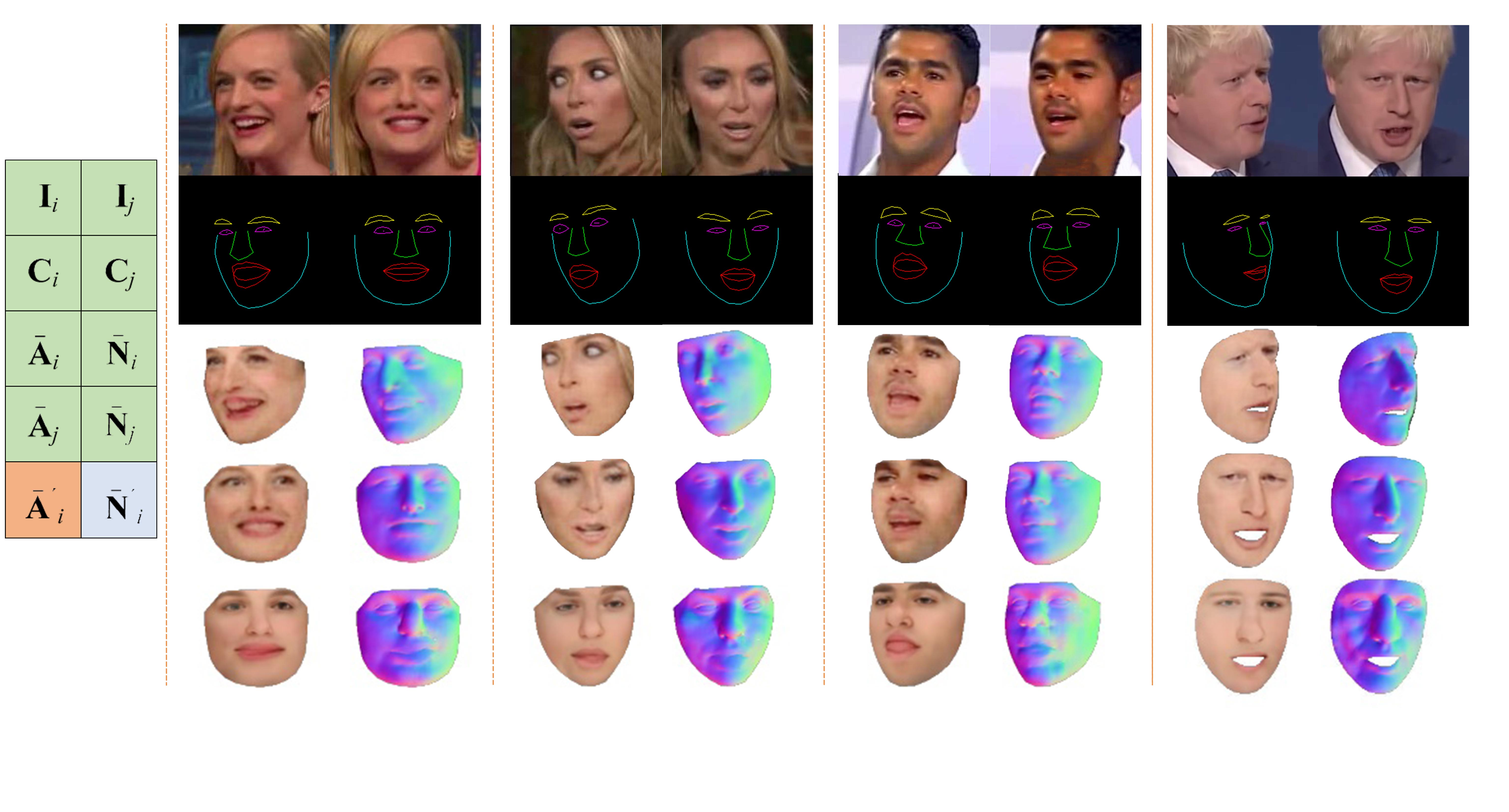}
\caption{Example outputs of $AlignNet_a$ and $AlignNet_n$ in IlluRes-SfSNet on real videos. }
\label{align_out} \end{figure}


Fig.~\ref{weakly_supervised_learning} shows our weakly supervised learning pipeline on real data. Here, we only introduce weakly supervised learning of \emph{AlignNet$_a$} based on consistency of albedo as 
the process for \emph{AlignNet$_n$} on normal is very similar. 
Consider two frames $\mathbf{I}_i$ and $\mathbf{I}_j$ from the video of a particular identity, our pre-trained IlluRes-SfSNet first predicts albedo, normal, lighting, and residual map for each of them.
By taking the predicted albedo $\bar{\mathbf{A}}_i$ of the $i$-th frame, together with contours $\mathbf{C}_i$ and $\mathbf{C}_j$ of the $i$-th and $j$-th frame, respectively, as inputs, \emph{AlignNet$_a$} first transforms $\bar{\mathbf{A}}_i$  to $\bar{\mathbf{A}}'_i$. Ideally, $\bar{\mathbf{A}}'_i$ and  albedo $\bar{\mathbf{A}}_j$ of the $j$-th frame should be identical under the consistency constraint.

We thus train \emph{AlignNet$_a$} using the loss function
\begin{equation}
L_{a} = \|\bar{\mathbf{A}}_j - \phi_a(\mathbf{C}_i, \mathbf{C}_j, \bar{\mathbf{A}}_i) \|,
\label{eq:loss_align_albedo}
\end{equation}
where $\phi_a$ transforms $\bar{\mathbf{A}}_i$ to $\bar{\mathbf{A}}'_i$ by applying the transformation between $\mathbf{C}_i$ and $\mathbf{C}_j$ on $\bar{\mathbf{A}}_i$.
Similarly, \emph{AlignNet$_n$} is trained with the loss function
\begin{equation}
L_{n} = \|\bar{\mathbf{N}}_j - \phi_n(\mathbf{C}_i, \mathbf{C}_j, \bar{\mathbf{N}}_i) \|,
\label{eq:loss_align_normal}
\end{equation}
where $\phi_n$ transforms $\bar{\mathbf{N}}_i$ to $\bar{\mathbf{N}}'_i$ by applying the transformation between $\mathbf{C}_i$ and $\mathbf{C}_j$ on $\bar{\mathbf{N}}_i$.

In this way, IlluRes-SfSNet can then be jointly trained with \emph{AlignNet$_a$} and  \emph{AlignNet$_n$} on real face videos by minimizing
\begin{equation}
L = \lambda_g\|\mathbf{I}_g - \bar{\mathbf{I}}_g \| + \lambda_{ab}L_a + \lambda_{no}L_n,
\label{loss_all}
\end{equation}
where $\lambda_{ab}$ and $\lambda_{no}$ are weights for \emph{AlignNet$_a$} and \emph{AlignNet$_n$} losses.
In practice, we first pre-train \emph{AlignNet$_a$} and \emph{AlignNet$_n$} using synthetic data with ground-truth $\mathbf{A}_{\{i, j\}}$ and  $\mathbf{N}_{\{i, j\}}$, and then conduct weakly supervised learning on real face videos.
Note that we align albedo and normal maps on two directions, namely, from $i$ to $j$ and from $j$ to $i$, in order to have stronger consistency.

We present a few example outputs of \emph{AlignNet$_{\{a,n\}}$} in Fig. \ref{align_out}. Note that although the prediction of \emph{AlignNet$_{\{a,n\}}$} is not perfect, our experiments verify that the consistency constraint on real face video can still improve the inverse rendering performance, which corroborates our assumption that coarse alignment on albedo and normal is enough to perform weakly supervised learning on real face images.

%% file: experiment.tex
\section{Experimental results}

\subsection{Datasets and implementation details}

\noindent \textbf{Synthetic data:} To conduct supervised training for our model, we make use of three synthetic datasets. Among them, one is an existing dataset and the other two are created by us based on other datasets. We first use the synthetic dataset provided by SfSNet \cite{sengupta2018sfsnet}, denoted as \emph{SfSNet-syn}. This dataset contains 12,500 identities. 15 images are rendered for each identity under varying illuminations and poses. The images are rendered by fitting the 3DMM \cite{blanz1999morphable}. In total, there are 187,500 images in this dataset. However, this dataset does not take expression variations and global illuminations into consideration.
Therefore we create two other synthetic datasets \emph{Expre-syn} and \emph{Caric-syn} to compensate for this.
For \emph{Expre-syn}, we randomly choose 114 video clips from VoxCeleb \cite{nagrani2017voxceleb} and each clip corresponds to one identity.
We fit the Basel Face Model \cite{paysan20093d} for each frame to approximate realistic poses and expressions under four random viewpoints and five random illuminations using Mitsuba \cite{Mitsuba}.
Six different expressions are fitted for each identity.
In total we obtain 13,680 images in \emph{Expre-syn}.
Since in practice there can be some exaggerated expressions that do not appear in VoxCeleb, we further create \emph{Caric-syn} to increase expression variations with 100 virtual identities.
For each identity, we render four distinct poses with six caricature expressions under five random lighting conditions, resulting in 12,000 synthetic face images.
Our data generation process inherently considers global illumination.
For each face image $\mathbf{I}_g$, we also obtain the corresponding albedo $\mathbf{A}$, normal $\mathbf{N}$, lighting $\mathbf{L}$, residual map $\mathbf{R}$, and face image $\mathbf{I}_l$ with local illumination.
A summary of the synthetic data is given in Table~\ref{tab:syn_data}.
We use images for one identity from each of \emph{Expre-syn} and \emph{Caric-syn} for testing and all the rest together with \emph{SfSNet-syn} for training.

 In addition, to train AlignNet, we further aggregate albedo and normal in \emph{Expre-syn} and \emph{Caric-syn}. Specifically, we add various textures into synthetic albedos to help $AlignNet_a$ focus on learning displacements.
 We create 50 texture styles and randomly transfer them onto face albedos.
 We also perform shape deformation on synthetic face models to avoid strong face priors in $AlignNet_n$.
 We design 20 different deformations on faces and build a linear combination basis. The deformations are generated using a sketch-based modelling system \cite{han2017deepsketch2face}.
 In total, we have 540 different albedos under four viewpoints and six expressions for $AlignNet_a$ training, and 540 identities with four viewpoints and six deformation styles (produced by randomly weighting the linear deformation bases) for $AlignNet_n$.
 Note that we only use this dataset to train AlignNet.

\noindent \textbf{Real data:}
We extract 114 real video clips from the 300VW dataset \cite{shen2015first}.
Each clip corresponds to one identity (with four viewpoints and five illuminations).
There are 3,462 images in total.
This dataset contains rich but non-exaggerated expressions.
 It is used for weakly supervised training only.
 For qualitative evaluations, we use face images in CelebA \cite{liu2018large} by picking 500 images with various face attributes(e.g., bread, glasses, and age).

 For quantitative evaluation on normal recovery, we adopt Photoface \cite{zafeiriou2011photoface}, which is created under various harsh illumination conditions and captures the ground truth of face normal.

\noindent \textbf{Implementation details:} Our network architecture is implemented in TensorFlow. The input images are all of size 128$\times$128.
We train our model using a batch size of four using the learning rate of 5e-4 with a decay rate of 0.98 and a decay step of five.
As a preparation procedure, we pre-train $AlignNet_a$ and $AlignNet_n$ with synthetic data, with a learning rate of 5e-4 for 30 epochs using a batch size of eight.
We then train our inverse rendering model in three stages: first, we train our IlluRes-SfSNet for 20 epochs on the combined synthetic dataset of SfSNet-syn, \emph{Caric-syn} , and \emph{Expre-syn}; second, we train IlluRes-SfSNet on real data by using the pre-trained $AlignNet_a$ and $AlignNet_n$ to provide weak supervision for 20 epochs; and third, we jointly fine-tune the entire network that includes IlluRes-SfSNet, $AlignNet_a$ and $AlignNet_n$ on both real and synthetic data.
During the first stage of training, for each epoch we randomly sample 30,000
images from \emph{SfSNet-syn} and use all the training data from \emph{Caric-syn} (13,560 images) and \emph{Expre-syn} (11,880 images) to avoid the training being biased by \emph{SfSNet-syn}.

 During the second stage of training, we use the real videos containing 3,487 images, \emph{Caric-syn}, and \emph{Expre-syn} on the fixed $AlignNet_{\{a, n\}}$ to train IlluRes-SfSNet. In each epoch, we randomly sample 30,000 pairs of real images and 16,272  pairs of \emph{Expre-syn} images.
 At the last stage, we use synthetic and real data to fine tune the entire network.

\begin{table}[h]
\centering
\footnotesize
\caption{Dataset construction with controlled variations for each set. Note that for \emph{SfSNet-syn} dataset, each identity has 15 images rendered under varying illuminations and poses.}
\begin{tabular}{|c|c|c|c|c|c|c|}
\hline
\multirow{2}{*}{Dataset} & \multirow{2}{*}{Iden.} & \multicolumn{4}{c|}{Per Identity}                                         & \multicolumn{1}{c|}{\multirow{2}{*}{\textbf{Total}}} \\ \cline{3-6}
                         &                        & Illu.            & Pose             & Expr.            & Exag.            & \multicolumn{1}{c|}{}                                \\ \hline
\emph{SfSNet-syn}                 & 12.5k                  & 15               & 15               & \textbackslash{} & \textbackslash{} & \textbf{187.5k}                                      \\ \cline{1-7}
\emph{Expre-syn}                & 114                    & 5                & 4                & 6 & \textbackslash{}                & \textbf{13.68k}                                       \\ \cline{1-7}
\emph{Caric-syn}                 & 100                   & 5                & 4                & \textbackslash{}                & 6 & \textbf{12.00k}                                       \\ \cline{1-7}
Real                     & 114                    & \textbackslash{} & \textbackslash{} & 25$\sim$35       & \textbackslash{} & \textbf{3462}                                        \\ \cline{1-7}
\end{tabular}
\label{tab:syn_data}
\end{table}

\subsection{Comparison with state of the art}\label{4.2}

\begin{figure*} [htb] \centering
\fontsize{16}{16}\selectfont
\includegraphics[width=4.8in]{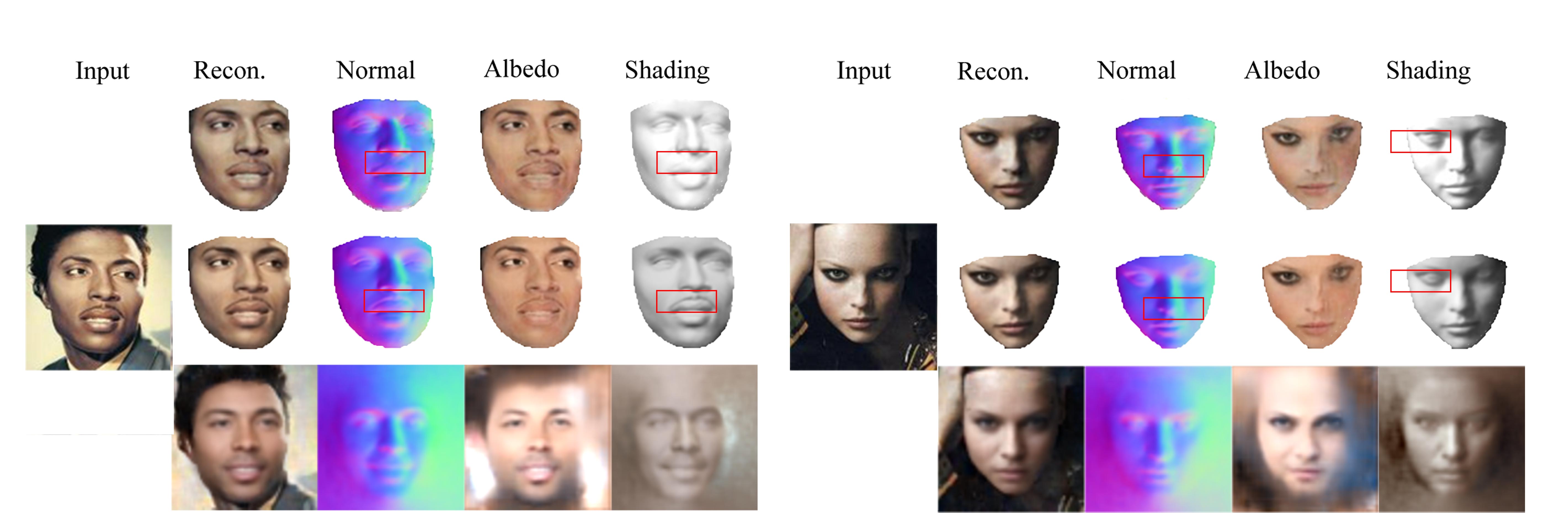}
\caption{Inverse rendering results on two samples. The first row of each sample shows our decompostion results. The following rows come from SfSNet\cite{sengupta2018sfsnet} and NeuralFace\cite{shu2017neural}. (Best viewed in PDF with zoom.)}
\label{comp_ir1} \end{figure*}

\noindent \textbf{Inverse rendering:} We compare our full model, IlluRes-SfSNet-Align, with NeuralFace\cite{shu2017neural} and SfSNet\cite{sengupta2018sfsnet}. Note that since different strategies are used for learning from real data in
\cite{shu2017neural}\cite{sengupta2018sfsnet}, we directly use their released pre-trained models to reproduce results
for inclusion in Fig.\ref{comp_ir1}.
It can be seen that IlluRes-SfSNet-Align and SfSNet generate more realistic decomposition than NeuralFace.
The regions in red rectangles show our method could alleviate the ambiguity on face shape, which are caused by makeup, such as the moustache in the first sample and the eyeliner in the second sample. Our method decomposes them correctly.
Moreover, compared with SfSNet, our model provides extra discernible shadows on albedo while
capturing more fine-scale details in the normal map.

\begin{figure*} [htb] \centering
\fontsize{8}{10}\selectfont
\includegraphics[width=4.8in]{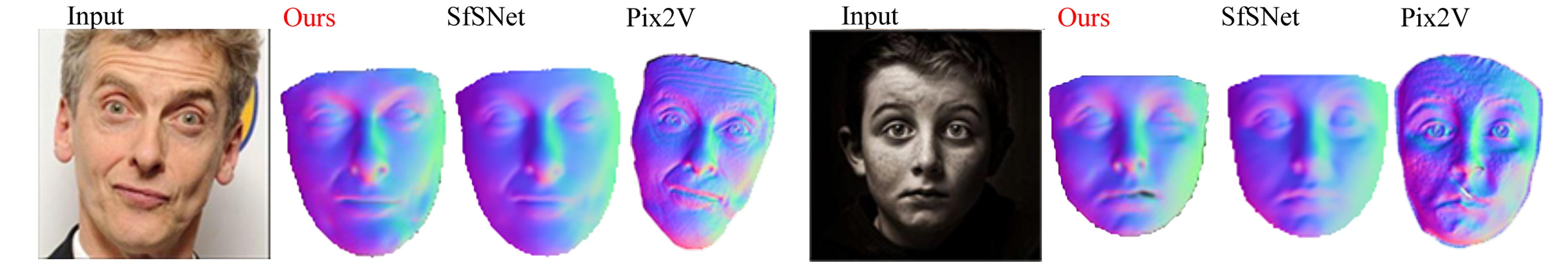}
\caption{Normal recovery results on the showcases of SfSNet\cite{sengupta2018sfsnet}. The first two columns are results from inverse rendering while the last column is from face reconstruction algorithm. }
\label{comp_nor1}

\end{figure*}

\noindent \textbf{Shape estimation:} Next we compare  the quality of shape estimation among lluRes-SfSNet-Align, 3DMM and Pix2Vertex \cite{sela2017unrestricted}. 
The dataset used for quantitative testing is Photoface \cite{zafeiriou2011photoface}, which provides ground truth of face normal maps under various illumination conditions. Following \cite{trigeorgis2017face}\cite{sengupta2018sfsnet}, we randomly pick 100 identities from a total of 454. We evaluate the performance by mean angular error of the normals and the percentage of pixels at various angular error thresholds and report them in Table \ref{tab:shape_compare}. The higher percentage of the forth column means a model could capture more low frequency shape information, while data of smaller angle measures the performance on estimating fine-scale structure. It can be seen that
3DMM achieves the highest percentage in the forth column but quite small value at the second one, which means it captures most low frequency shape while losing lots of detailed information.
Pix2Vertex captures much more details in the images but fails to recover shape information from harsh lighting and unusual expression, as the lowest value in the forth column suggested. 
Fig.\ref{comp_nor1} compares estimated normals of the showcases in \cite{sengupta2018sfsnet}.  
The normal images agrees with the Table \ref{tab:shape_compare} that our model outperforms SfSNet on normal estimation.

\begin{table}[h]
\centering
\small
\caption{Comparison on normal reconstruction error on the Photoface dataset. Lower is better for column 1, and higher is better for the percentage of pixels at specific thresholds. }
\begin{tabular}{|c|c|c|c|c|}
\hline
Algorithm  & Mean $\pm$ std  & \textless{}\ang{20} & \textless{}\ang{25} & \textless{}\ang{30} \\ \hline
3DMM       &31.9 $\pm$ 11.6  & 3.7\%         & 53.2\%        & 87.3\%        \\ \hline
Pix2Vertex\cite{sela2017unrestricted} &36.3 $\pm$ 6.2   & 23.6\%        & 35.4\%        & 46.1\%        \\ \hline
SfSNet\cite{sengupta2018sfsnet}     &   30.7$\pm$ 8.6         & 40.5\%        & 54.2\%        & 64.5\%        \\ \hline
Ours       &$\bm{25.3\pm 6.3}$            & $\bm{43.6}$\%        & $\bm{57.6}$\%        & $\bm{68.8}$\%        \\ \hline
\end{tabular}

\label{tab:shape_compare}
\end{table}

\subsection{Ablation Studies}\label{4.3}

\noindent \textbf{IlluRes-SfSNet v. SfSNet:}
To demonstrate the effectiveness of weakly supervised learning in Illuses-SfSNet, we use a total of five models. They are

\noindent \textbf{SfS-pretrain:} The authors of \cite{sengupta2018sfsnet} provide this model which is trained by using ‘SfS-supervision’. They first use a `skip-net' to obtain \textbf{A} and \textbf{N} of real images, then combine synthetic and real data with `ground-truth' to train SfSNet.
As SfS-pretrain has been trained with `SfS-supervision', real images can be reconstructed. However, SfSNet's structure cannot handle global illumination effects regardless of how training is done. 
These effects always exist on albedo and/or normal.

\noindent \textbf{SfS:} We use synthetic data to train SfS. In some situations this model outputs reasonable normal or albedo maps. Since it cannot be trained on real images, its overall reconstruction results are not satisfactory.

\noindent \textbf{SfS-Align:} Adding \emph{AlignNet} to allow weakly supervised training in SfS. 
This model can effectively decompose and reconstruct real images. Its performance is similar to that of SfS-pretrain.

\noindent \textbf{Illuses-SfS-pre:} We train Illures-SfS with synthetic data without using \emph{AlignNet}. IlluRes-SfS-pre successfully compensates for defects in SfS by separating the residual portion. Without weakly supervised learning, this network cannot bridge the gap between real and synthetic data, hence its reconstruction results are not competitive.

\noindent \textbf{IlluRes-SfSNet:} Adding \emph{AlignNet} to Illuses-SfS-pre. This model separates the residual through IlluResNet to capture fine image details. Moreover, \emph{AlignNet} enables the network to effectively decompose and reconstruct real images.

\begin{figure*} [htb] \centering
\fontsize{6}{10}\selectfont
\includegraphics[width=4.8in]{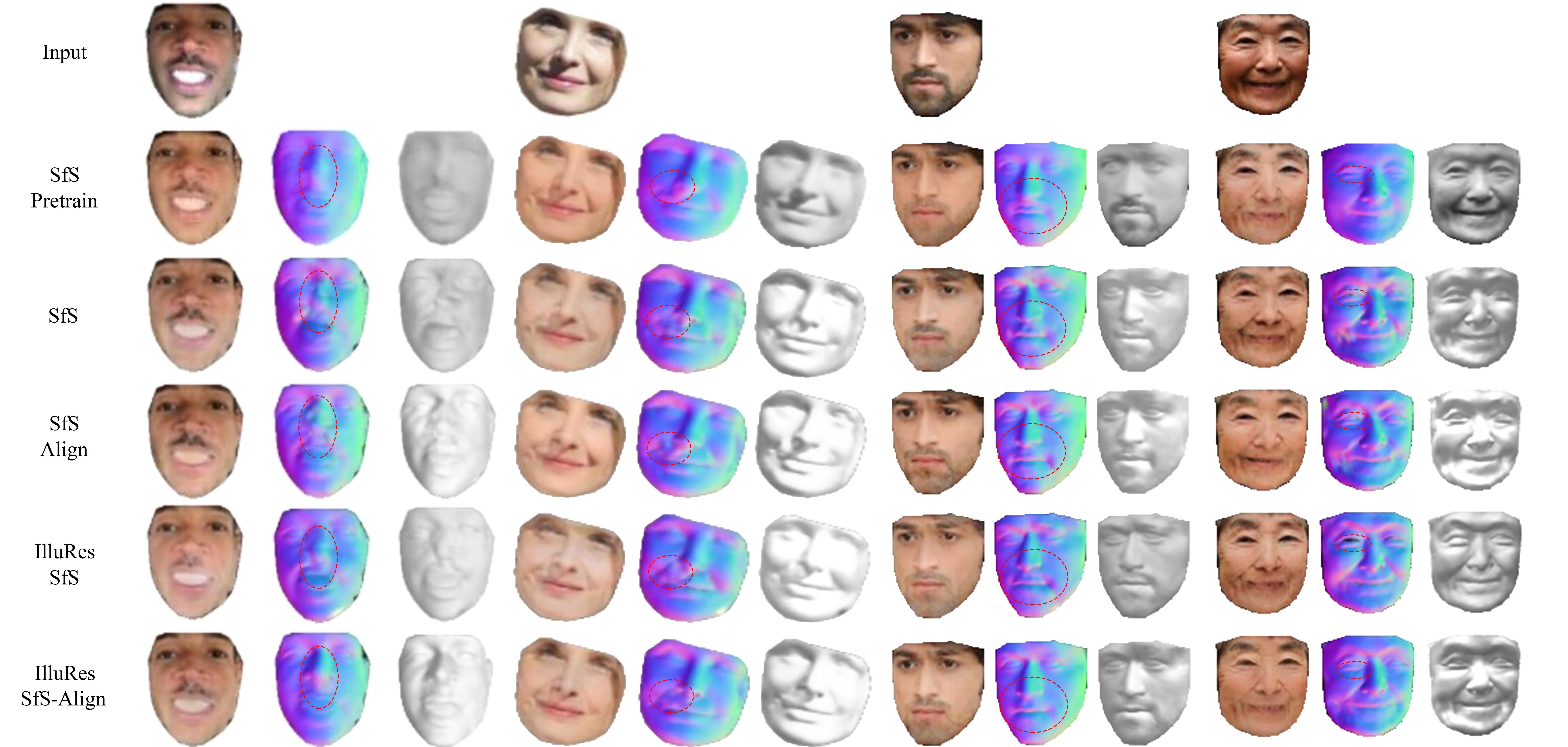}
\caption{Five models and four examples prove our \emph{AlignNet} and IlluResNet are effective. (Best viewed in PDF with zoom.)}
\label{fig_compare_five_net} \end{figure*}

Fig.\ref{fig_compare_five_net} compares decomposition and reconstruction results of the above five models. In the comparison of decomposition, our model is optimal, which proves that our \emph{AlignNet} and Residual net are effective.

%% file: application.tex
\subsection{Application}
Based on the inverse rendering results of our method, we are able to develop a wide range of applications. Here we show two of examples, namely, face relighting and albedo editing.

\begin{figure} [htb] \centering
\fontsize{8}{10}\selectfont
\includegraphics[width=4.5in]{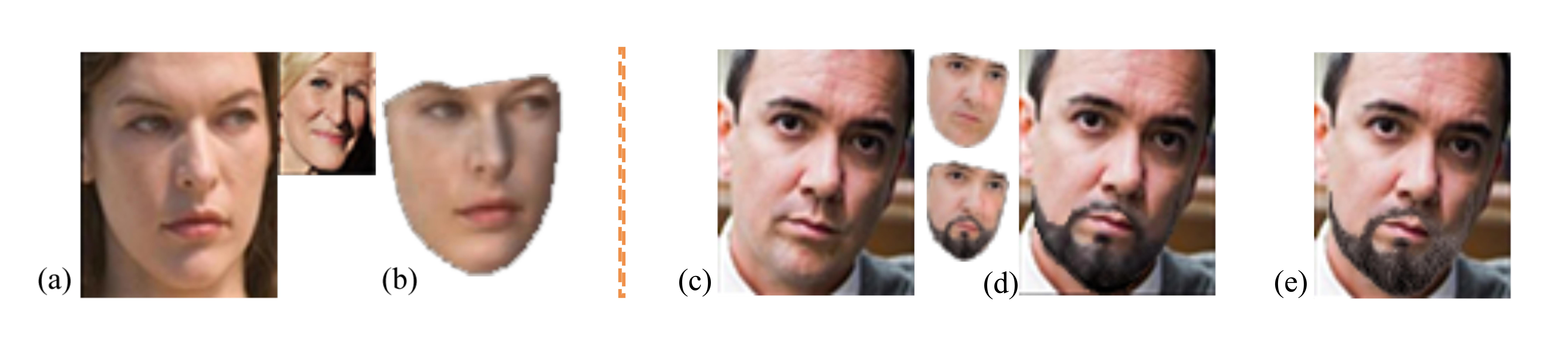}
\caption{The left of the figure is case of photo relighting. (a) are the source images and (b) are the results corresponding to the light information in the small photos.The right of the figure is case of albedo editing. (c) input photo, (d) result of albedo editing, (e) result of face photo editing. The corresponding albedo and modified one are shown in the small images.}
\label{relit}
\end{figure}

To relight photo, source and target images are sent into the well-trained IlluRes-SfSNet-Align, decomposed into corresponding albedo, normal, and lighting information. The lighting of source image is replaced by the target one and the novel combination of components are used to generate the relighted image. Fig.\ref{relit} shows the results.


Similarly, in albedo editing, the source photo is fed into our full model and produces the decomposition components. After modifying the albedo map, a new photo could be generated. We demonstrate a sample of beard editing in Fig.\ref{relit}. The result shows editing on albedo could generate more realistic result than directly editing on face photo. For example, we can see that the cheek in Fig.\ref{relit}(e) contains obvious artifacts, while the result of our model in Fig.\ref{relit}(d) is much more visually pleasing.

%% file: conclusion.tex
\section{Conclusion}
In this paper we propose a weakly supervised approach for inverse face rendering on real face videos, based on the assumption of consistency of albedo and normal of the same person in a video, bridging the domain gap between synthetic and real face. We propose \emph{AlignNet$_{\{a,n\}}$} to align albedo and normal subspaces between different frames of a certain video clip. We empirically show that the alignment (though not perfect) obtained by \emph{AlignNet$_{\{a,n\}}$} can provide enough constraints on frame consistency for weakly supervised learning on real images. Together with IlluRes-SfSNet, our framework has strong capability to disentangle normal and albedo into separate subspaces. Qualitative and quantitative evaluations show that our method outperforms state-of-the-art works on face inverse rendering.

%% file: Appendix.tex
\section*{Appendix}
\renewcommand\thesubsection{\Alph{subsection}}
\subsection{Architectures of AlignNet and IlluResNet}
In this section, we describe our network in detail. To better illustrate, we name layers in the figures with abbreviations: $C$ as Convolution layer followed by Batch Normalization, $CT$ as Transposed Convolution layer followed by Batch Normalization, $AP$ as Average Pooling layer and $Res$ as Residual Block. Each $Res$ consists of BN - ReLU - C128 - BN - ReLU - C128. For net parameters, $k$ represents kernal size and $s$ is stride. The parameter $u$ of CT means the size of output after upsampling. $LR$ is the abbreviation of Leaky ReLU and $R$ is ReLU. 

\begin{figure}[htb]
  \centering
  \includegraphics[width=4.2in]{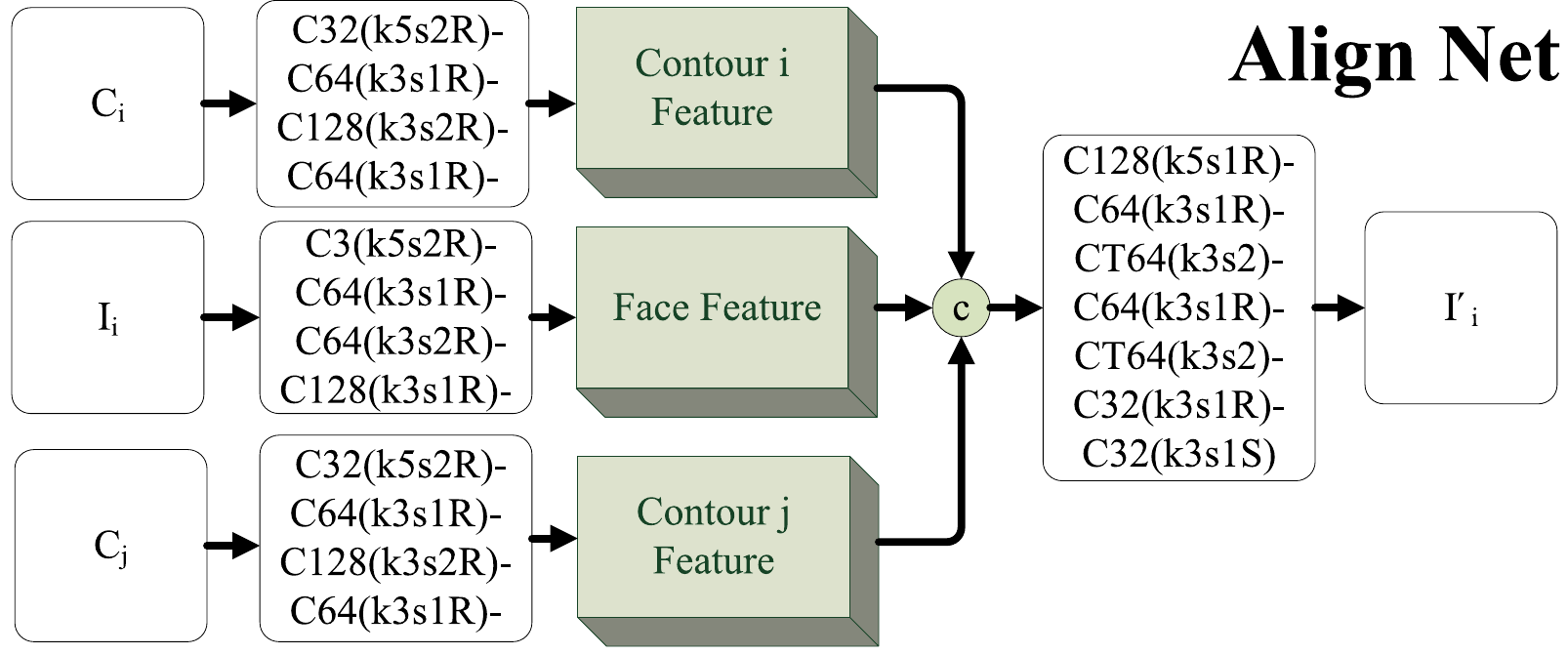}
  \caption{AlignNet Architecture.}
  \label{align}
\end{figure}

We first propose AlignNet for albedo and normal, shown in Fig.\ref{align}. $\mathbf{I}_i$ is albedo or normal of source frame and $\mathbf{I}'_i$ is the estimated component of target frame. $\mathbf{C}_i$ and $\mathbf{C}_j$ are the corresponding face contour of source frame $i$ and target frame $j$. These two networks have the same structure while they are trained on different data. 

\begin{figure}[htb]
  \centering
  \includegraphics[width=4.2in]{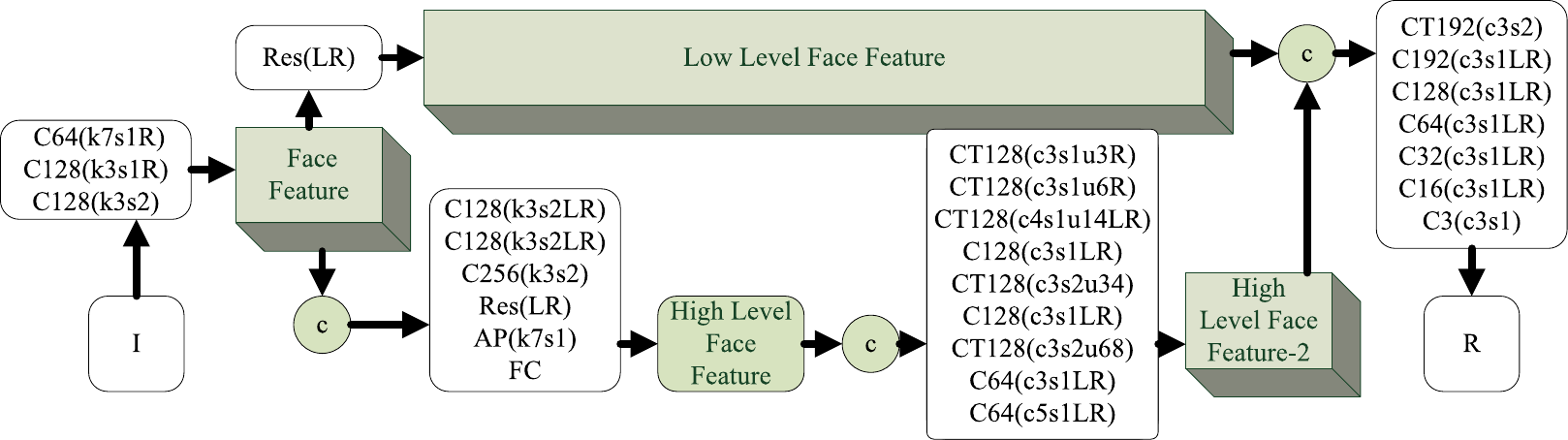}
  \caption{IlluResNet Architecture.}
  \label{IlluresNet}
\end{figure}

Details of IlluResNet are provided in Fig.\ref{IlluresNet}.
\subsection{More Qualitative Results}
\begin{figure*}[tb]
  \centering
  \includegraphics[width=3.8in]{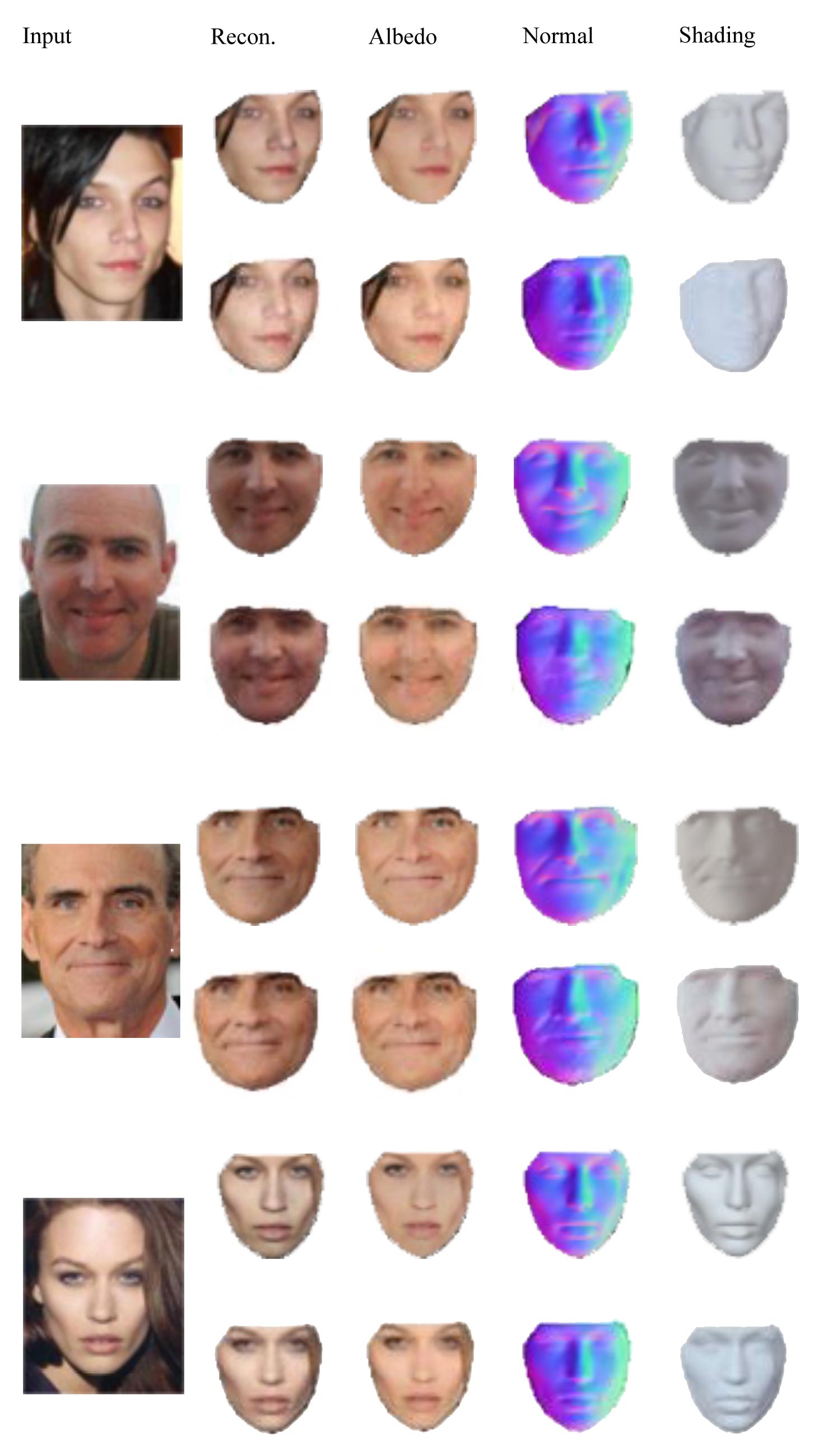}
  \caption{Inverse rendering. The first row of each sample is from IlluRes-SfSNet-Align, while the second one from SfSNet. In general, our method captures more details on normal, such as contour of nose and eyes, and winkles on face.}
  \label{sfs_1}
\end{figure*}

\begin{figure*}[tb]
  \centering
  \includegraphics[width=3.8in]{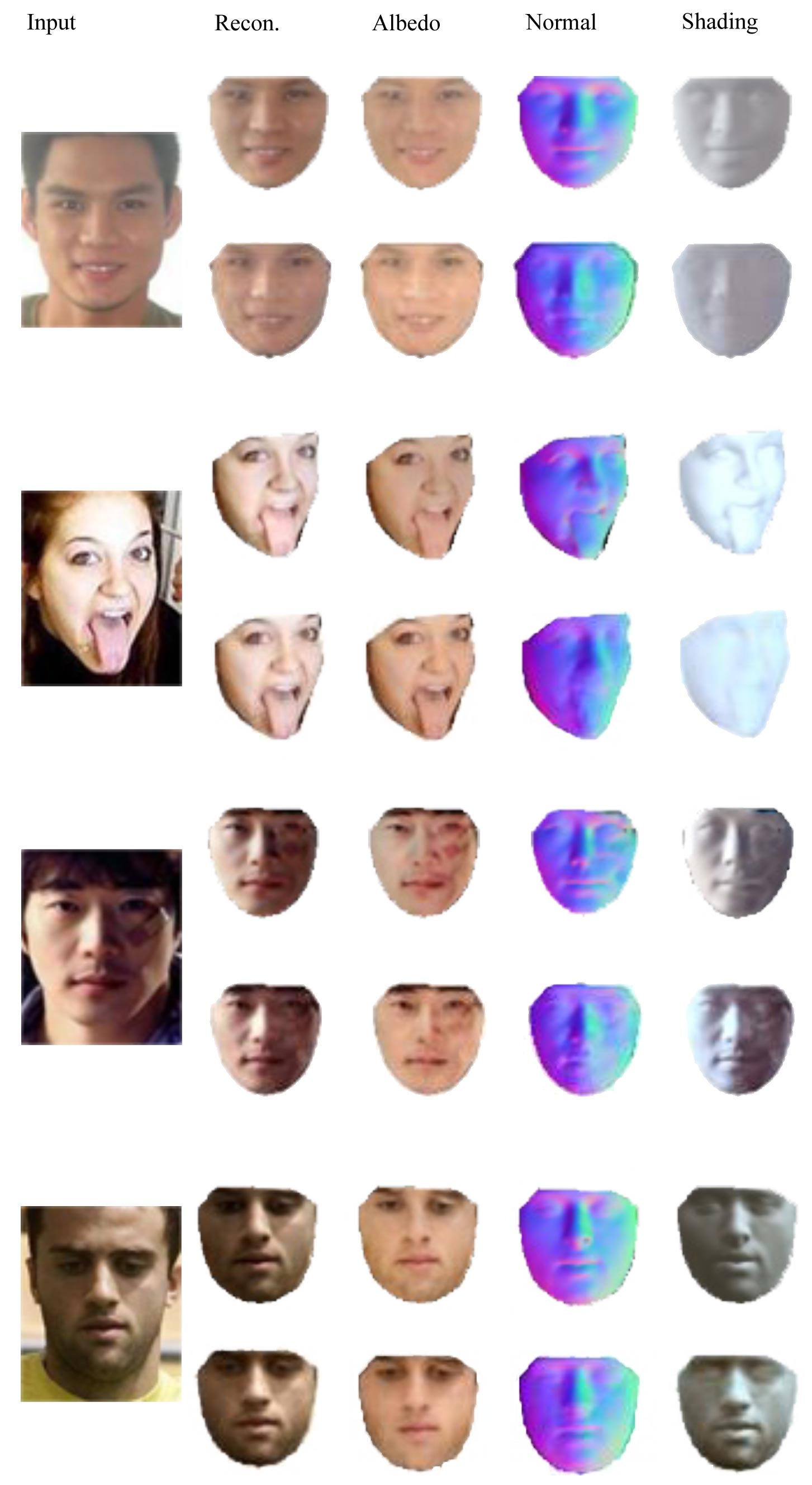}
  \caption{Inverse rendering. The first row of each sample is from IlluRes-SfSNet-Align, while the second one from SfSNet. }
  \label{sfs_2}
\end{figure*}

In Fig.\ref{sfs_1} and Fig. \ref{sfs_2}, we present inverse rendering results from our full model,  IlluRes-SfSNet-Align, and SfSNet. The input images are sampled from CelebA\cite{liu2018large}. 

\begin{figure*}[tb]
  \centering
  \includegraphics[width=4in]{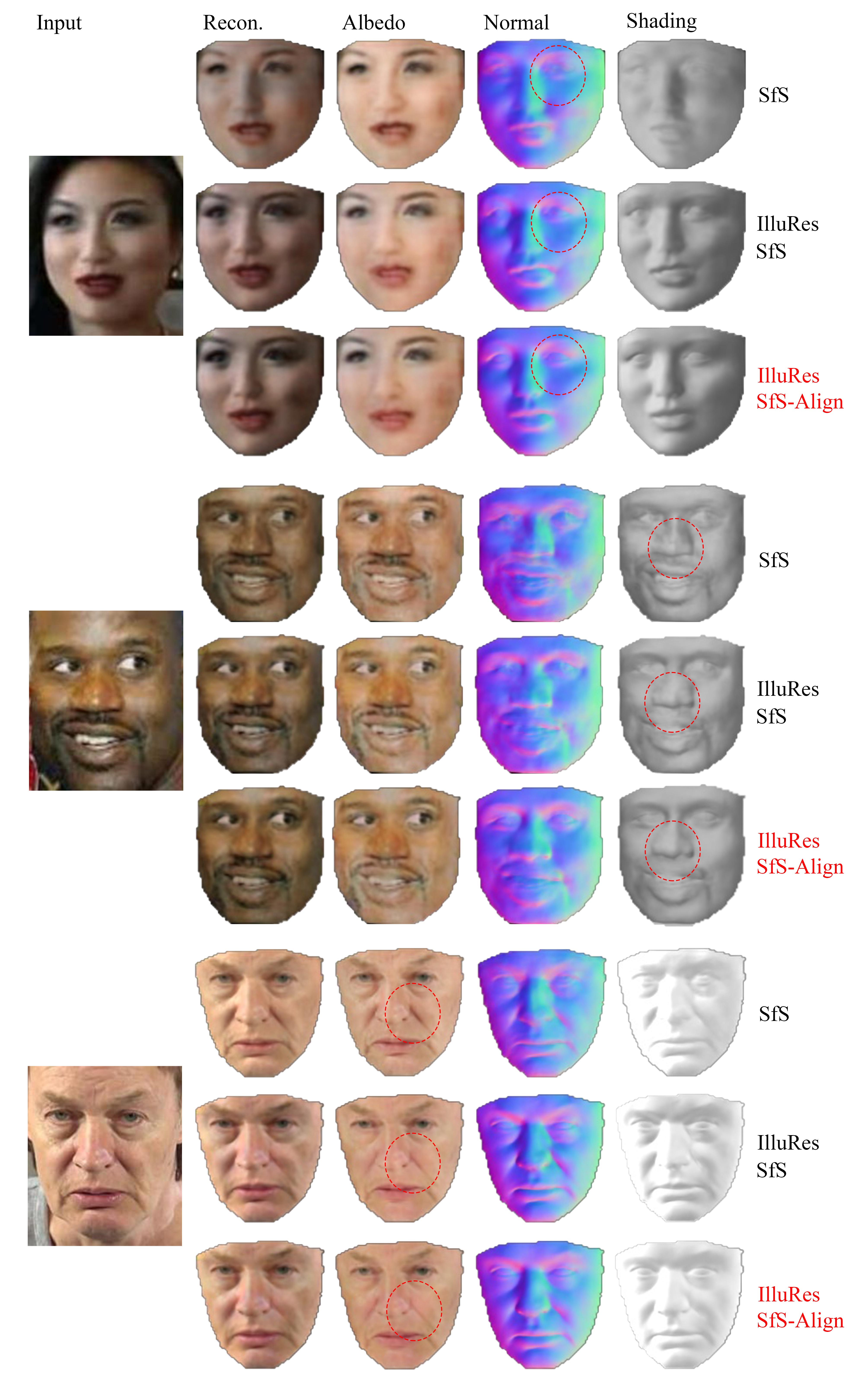}
  \caption{Inverse rendering at $256 \times 256$ resolution. The red circles highlight our improvements. }
  \label{256_1}
\end{figure*}

In Fig.\ref{256_1}, we further compare our method against SfSNet with a higher input resolution of $256 \times 256$ (all the two networks are re-trained for such a new setting). 
With a higher resolution, it can be seen that our method can retain the details much better than SfSNet.

\begin{figure*}[tb]
  \small
  \centering
  \includegraphics[width=4.2in]{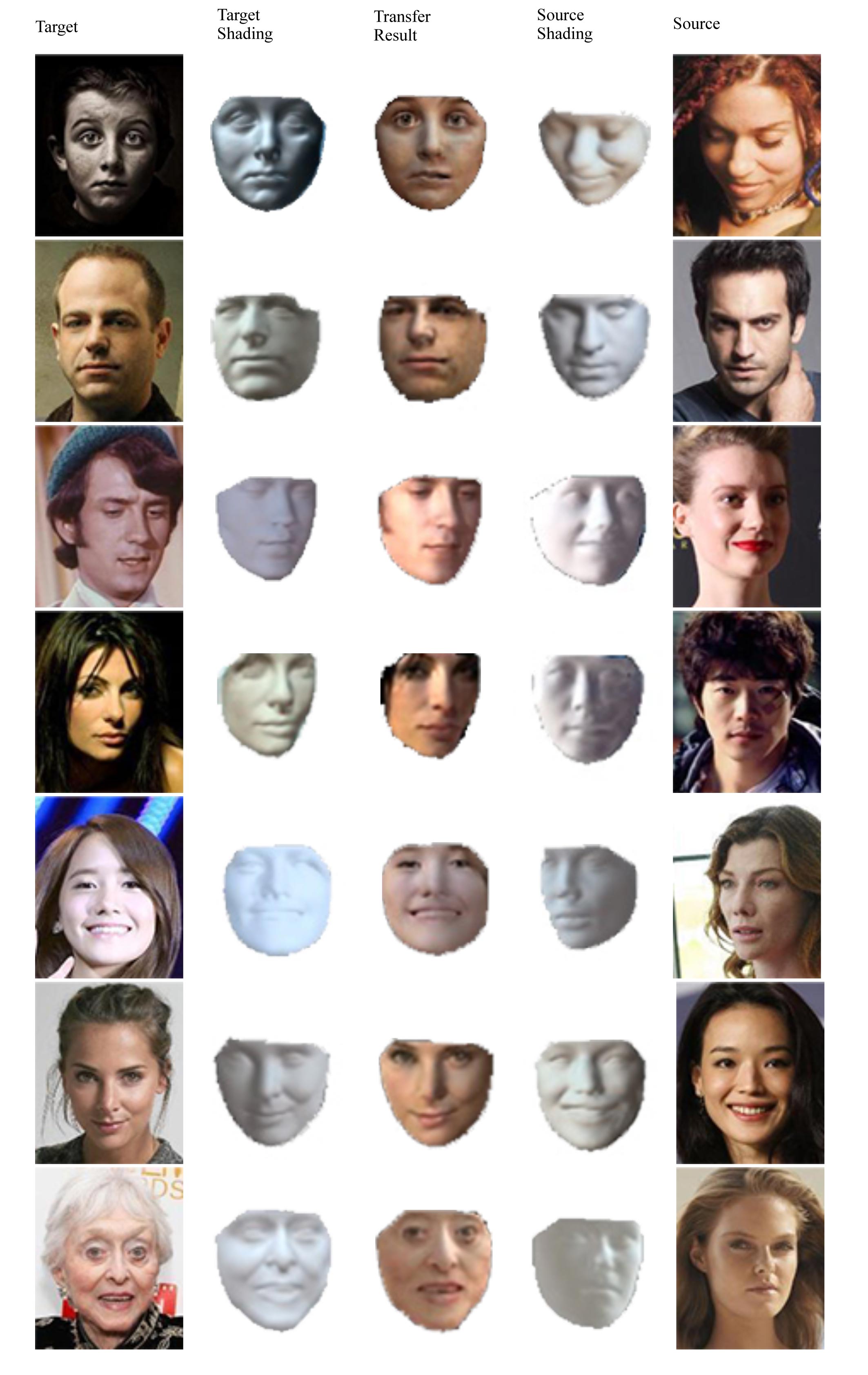}
  \caption{Lighting transfer. Our method can transfer the lighting condition in source photo to target photo.}
  \label{relit_1}
\end{figure*}

In Fig.\ref{relit_1}, we provide some samples for lighting transfer on faces.

%% file: eccv2020submission.bbl
\begin{thebibliography}{10}
\providecommand{\url}[1]{\texttt{#1}}
\providecommand{\urlprefix}{URL }
\providecommand{\doi}[1]{https://doi.org/#1}

\bibitem{Faceplusplus}
Faceplusplus. \url{www.faceplusplus.com} (cited Nov 2019)

\bibitem{aldrian2012inverse}
Aldrian, O., Smith, W.A.: Inverse rendering of faces with a 3d morphable model.
  IEEE transactions on pattern analysis and machine intelligence
  \textbf{35}(5),  1080--1093 (2012)

\bibitem{barron2013intrinsic}
Barron, J.T., Malik, J.: Intrinsic scene properties from a single rgb-d image.
  In: Proceedings of the IEEE Conference on Computer Vision and Pattern
  Recognition. pp. 17--24 (2013)

\bibitem{barron2014shape}
Barron, J.T., Malik, J.: Shape, illumination, and reflectance from shading.
  IEEE transactions on pattern analysis and machine intelligence
  \textbf{37}(8),  1670--1687 (2014)

\bibitem{barrow1978recovering}
Barrow, H.: Recovering intrinsic scene characteristics  (1978)

\bibitem{blanz1999morphable}
Blanz, V., Vetter, T., et~al.: A morphable model for the synthesis of 3d faces.
  (1999)

\bibitem{bonneel2017intrinsic}
Bonneel, N., Kovacs, B., Paris, S., Bala, K.: Intrinsic decompositions for
  image editing. In: Computer Graphics Forum. vol.~36, pp. 593--609. Wiley
  Online Library (2017)

\bibitem{bousseau2009user}
Bousseau, A., Paris, S., Durand, F.: User-assisted intrinsic images. In: ACM
  Transactions on Graphics (TOG). vol.~28, p.~130. ACM (2009)

\bibitem{bradley2010high}
Bradley, D., Heidrich, W., Popa, T., Sheffer, A.: High resolution passive
  facial performance capture. In: ACM transactions on graphics (TOG). vol.~29,
  p.~41. ACM (2010)

\bibitem{chen2013simple}
Chen, Q., Koltun, V.: A simple model for intrinsic image decomposition with
  depth cues. In: Proceedings of the IEEE International Conference on Computer
  Vision. pp. 241--248 (2013)

\bibitem{debevec2000acquiring}
Debevec, P., Hawkins, T., Tchou, C., Duiker, H.P., Sarokin, W., Sagar, M.:
  Acquiring the reflectance field of a human face. In: Proceedings of the 27th
  annual conference on Computer graphics and interactive techniques. pp.
  145--156. ACM Press/Addison-Wesley Publishing Co. (2000)

\bibitem{fan2018revisiting}
Fan, Q., Yang, J., Hua, G., Chen, B., Wipf, D.: Revisiting deep intrinsic image
  decompositions. In: Proceedings of the IEEE conference on computer vision and
  pattern recognition. pp. 8944--8952 (2018)

\bibitem{ghosh2011multiview}
Ghosh, A., Fyffe, G., Tunwattanapong, B., Busch, J., Yu, X., Debevec, P.:
  Multiview face capture using polarized spherical gradient illumination. In:
  ACM Transactions on Graphics (TOG). vol.~30, p.~129. ACM (2011)

\bibitem{han2017deepsketch2face}
Han, X., Gao, C., Yu, Y.: Deepsketch2face: a deep learning based sketching
  system for 3d face and caricature modeling. ACM Transactions on Graphics
  (TOG)  \textbf{36}(4), ~126 (2017)

\bibitem{horn1974determining}
Horn, B.K.: Determining lightness from an image. Computer graphics and image
  processing  \textbf{3}(4),  277--299 (1974)

\bibitem{Mitsuba}
Jakob, W.: Mitsuba renderer (2010), http://www.mitsuba-renderer.org

\bibitem{kemelmacher20103d}
Kemelmacher-Shlizerman, I., Basri, R.: 3d face reconstruction from a single
  image using a single reference face shape. IEEE transactions on pattern
  analysis and machine intelligence  \textbf{33}(2),  394--405 (2010)

\bibitem{Kim2016MultiviewIR}
Kim, K., Torii, A., Okutomi, M.: Multi-view inverse rendering under arbitrary
  illumination and albedo. In: ECCV (2016)

\bibitem{kimmel2003variational}
Kimmel, R., Elad, M., Shaked, D., Keshet, R., Sobel, I.: A variational
  framework for retinex. International Journal of computer vision
  \textbf{52}(1),  7--23 (2003)

\bibitem{laffont2015intrinsic}
Laffont, P.Y., Bazin, J.C.: Intrinsic decomposition of image sequences from
  local temporal variations. In: Proceedings of the IEEE International
  Conference on Computer Vision. pp. 433--441 (2015)

\bibitem{lee2005estimation}
Lee, J., Machiraju, R., Pfister, H., Moghaddam, B.: Estimation of 3d faces and
  illumination from single photographs using a bilineaur illumination model
  (2005)

\bibitem{lee2012estimation}
Lee, K.J., Zhao, Q., Tong, X., Gong, M., Izadi, S., Lee, S.U., Tan, P., Lin,
  S.: Estimation of intrinsic image sequences from image+ depth video. In:
  European Conference on Computer Vision. pp. 327--340. Springer (2012)

\bibitem{lettry2018darn}
Lettry, L., Vanhoey, K., Van~Gool, L.: Darn: a deep adversarial residual
  network for intrinsic image decomposition. In: 2018 IEEE Winter Conference on
  Applications of Computer Vision (WACV). pp. 1359--1367. IEEE (2018)

\bibitem{li2014intrinsic}
Li, C., Zhou, K., Lin, S.: Intrinsic face image decomposition with human face
  priors. In: European conference on computer vision. pp. 218--233. Springer
  (2014)

\bibitem{Li2018LearningII}
Li, Z., Snavely, N.: Learning intrinsic image decomposition from watching the
  world. 2018 IEEE/CVF Conference on Computer Vision and Pattern Recognition
  pp. 9039--9048 (2018)

\bibitem{liu2018large}
Liu, Z., Luo, P., Wang, X., Tang, X.: Large-scale celebfaces attributes
  (celeba) dataset

\bibitem{ma2018single}
Ma, W.C., Chu, H., Zhou, B., Urtasun, R., Torralba, A.: Single image intrinsic
  decomposition without a single intrinsic image. In: Proceedings of the
  European Conference on Computer Vision (ECCV). pp. 201--217 (2018)

\bibitem{nagrani2017voxceleb}
Nagrani, A., Chung, J.S., Zisserman, A.: Voxceleb: a large-scale speaker
  identification dataset. arXiv preprint arXiv:1706.08612  (2017)

\bibitem{narihira2015direct}
Narihira, T., Maire, M., Yu, S.X.: Direct intrinsics: Learning albedo-shading
  decomposition by convolutional regression. In: Proceedings of the IEEE
  international conference on computer vision. pp. 2992--2992 (2015)

\bibitem{paysan20093d}
Paysan, P., Knothe, R., Amberg, B., Romdhani, S., Vetter, T.: A 3d face model
  for pose and illumination invariant face recognition. In: 2009 Sixth IEEE
  International Conference on Advanced Video and Signal Based Surveillance. pp.
  296--301. Ieee (2009)

\bibitem{rother2011recovering}
Rother, C., Kiefel, M., Zhang, L., Sch{\"o}lkopf, B., Gehler, P.V.: Recovering
  intrinsic images with a global sparsity prior on reflectance. In: Advances in
  neural information processing systems. pp. 765--773 (2011)

\bibitem{schneider2017efficient}
Schneider, A., Schonborn, S., Frobeen, L., Egger, B., Vetter, T.: Efficient
  global illumination for morphable models. In: Proceedings of the IEEE
  International Conference on Computer Vision. pp. 3865--3873 (2017)

\bibitem{sela2017unrestricted}
Sela, M., Richardson, E., Kimmel, R.: Unrestricted facial geometry
  reconstruction using image-to-image translation. In: Proceedings of the IEEE
  International Conference on Computer Vision. pp. 1576--1585 (2017)

\bibitem{sengupta2018sfsnet}
Sengupta, S., Kanazawa, A., Castillo, C.D., Jacobs, D.W.: Sfsnet: Learning
  shape, reflectance and illuminance of facesin the wild'. In: Proceedings of
  the IEEE Conference on Computer Vision and Pattern Recognition. pp.
  6296--6305 (2018)

\bibitem{shen2013intrinsic}
Shen, J., Yang, X., Li, X., Jia, Y.: Intrinsic image decomposition using
  optimization and user scribbles. IEEE transactions on cybernetics
  \textbf{43}(2),  425--436 (2013)

\bibitem{shen2015first}
Shen, J., Zafeiriou, S., Chrysos, G.G., Kossaifi, J., Tzimiropoulos, G.,
  Pantic, M.: The first facial landmark tracking in-the-wild challenge:
  Benchmark and results. In: Proceedings of the IEEE International Conference
  on Computer Vision Workshops. pp. 50--58 (2015)

\bibitem{shen2011intrinsic}
Shen, L., Yeo, C.: Intrinsic images decomposition using a local and global
  sparse representation of reflectance. In: CVPR 2011. pp. 697--704. IEEE
  (2011)

\bibitem{shu2017neural}
Shu, Z., Yumer, E., Hadap, S., Sunkavalli, K., Shechtman, E., Samaras, D.:
  Neural face editing with intrinsic image disentangling. In: Proceedings of
  the IEEE Conference on Computer Vision and Pattern Recognition. pp.
  5541--5550 (2017)

\bibitem{tappen2003recovering}
Tappen, M.F., Freeman, W.T., Adelson, E.H.: Recovering intrinsic images from a
  single image. In: Advances in neural information processing systems. pp.
  1367--1374 (2003)

\bibitem{tewari2019fml}
Tewari, A., Bernard, F., Garrido, P., Bharaj, G., Elgharib, M., Seidel, H.P.,
  P{\'e}rez, P., Zollhofer, M., Theobalt, C.: Fml: face model learning from
  videos. In: Proceedings of the IEEE Conference on Computer Vision and Pattern
  Recognition. pp. 10812--10822 (2019)

\bibitem{tewari2017mofa}
Tewari, A., Zollhofer, M., Kim, H., Garrido, P., Bernard, F., Perez, P.,
  Theobalt, C.: Mofa: Model-based deep convolutional face autoencoder for
  unsupervised monocular reconstruction. In: Proceedings of the IEEE
  International Conference on Computer Vision. pp. 1274--1283 (2017)

\bibitem{trigeorgis2017face}
Trigeorgis, G., Snape, P., Kokkinos, I., Zafeiriou, S.: Face normals. In:
  Proceedings of the IEEE Conference on Computer Vision and Pattern
  Recognition. pp. 38--47 (2017)

\bibitem{wang2008face}
Wang, Y., Zhang, L., Liu, Z., Hua, G., Wen, Z., Zhang, Z., Samaras, D.: Face
  relighting from a single image under arbitrary unknown lighting conditions.
  IEEE Transactions on Pattern Analysis and Machine Intelligence
  \textbf{31}(11),  1968--1984 (2008)

\bibitem{yu2019inverserendernet}
Yu, Y., Smith, W.A.: Inverserendernet: Learning single image inverse rendering.
  In: Proceedings of the IEEE Conference on Computer Vision and Pattern
  Recognition. pp. 3155--3164 (2019)

\bibitem{zafeiriou2011photoface}
Zafeiriou, S., Hansen, M., Atkinson, G., Argyriou, V., Petrou, M., Smith, M.,
  Smith, L.: The photoface database. In: CVPR 2011 WORKSHOPS. pp. 132--139.
  IEEE (2011)

\bibitem{zakharov2019few}
Zakharov, E., Shysheya, A., Burkov, E., Lempitsky, V.: Few-shot adversarial
  learning of realistic neural talking head models. In: Proceedings of the IEEE
  International Conference on Computer Vision. pp. 9459--9468 (2019)

\end{thebibliography}
